\documentclass[10pt,twocolumn,letterpaper]{article}

\usepackage{cvpr}              %
\usepackage{booktabs}
\usepackage{pifont}
\usepackage{array}
\usepackage{todonotes}
\usepackage{lipsum}
\usepackage{multirow}
\usepackage{xcolor}
\usepackage{float}
\usepackage[table]{xcolor}
\usepackage[absolute,overlay]{textpos}
\usepackage[dvipsnames]{xcolor}
\newcommand{\cmark}{\ding{51}}
\newcommand{\xmark}{\ding{55}}
\newcommand{\figref}[1]{Fig.~\ref{#1}}
\newcommand{\secref}[1]{Section~\ref{#1}}

\newcommand{\tabref}[1]{Table~\ref{#1}}

\makeatletter
\DeclareRobustCommand\onedot{\futurelet\@let@token\@onedot}
\def\@onedot{\ifx\@let@token.\else.\fi}

\makeatother

\definecolor{darkgreen}{rgb}{0,0.7,0}
\definecolor{darkblue}{RGB}{31,119,180}
\definecolor{darkred}{RGB}{214,39,40}
\definecolor{mediumgray}{rgb}{0.5,0.5,0.5}
\definecolor{mediumteal}{rgb}{0,0.5,0.5}
\definecolor{naviblue}{RGB}{0,0,128}

\definecolor{amber}{HTML}{FFBE28}
\definecolor{flame}{HTML}{E65100}
\definecolor{indigo}{HTML}{1A237E}
\definecolor{cyan}{HTML}{00ACC1}

\definecolor{teal}{HTML}{006064}
\definecolor{fern}{HTML}{81C784}
\definecolor{mint}{HTML}{E8F5E9}
\definecolor{slate}{HTML}{263238}

\definecolor{ellisred}{rgb}{0.87,0.44,0.38} %
\definecolor{ellisgreen}{rgb}{0.69,0.90,0.52} %
\definecolor{elliscyan}{rgb}{0.29,0.77,0.74} %
\definecolor{ellisorange}{rgb}{0.89,0.55,0.28} %
\definecolor{ellisblue}{rgb}{0.41,0.61,0.86} %

\definecolor{customgray}{RGB}{136, 138, 133}

\newcommand{\boldparagraph}[1]{\vspace{0.1cm}\noindent{\bf #1:} }

\definecolor{darkgreen}{rgb}{0,0.7,0}
\definecolor{lgray}{rgb}{0.6,0.6,0.6}

\definecolor{mediumgray}{rgb}{0.5,0.5,0.5}
\newcommand{\pmsd}[1]{{\color{mediumgray}{\scriptsize $\pm$ #1}}}

\definecolor{cvprblue}{rgb}{0.21,0.49,0.74}
\usepackage[pagebackref,breaklinks,colorlinks,allcolors=cvprblue,urlcolor=cvprblue]{hyperref}

\def\paperID{513} %
\def\confName{CVPR}
\def\confYear{2026}

\hypersetup{
  colorlinks=true,
  urlcolor=flame, 
  linkcolor=flame,
  citecolor=indigo,
  pdfborder={0 0 0},
}

\title{LEAD: Minimizing Learner–Expert Asymmetry in End-to-End Driving}

\author{
Long Nguyen{$^{1,3}$} \qquad
Micha Fauth{$^{1}$} \qquad
Bernhard Jaeger{$^{1,3}$} \qquad
Daniel Dauner{$^{1,3}$} \\
Maximilian Igl{$^{2}$} \qquad
Andreas Geiger{$^{1,3}$} \qquad
Kashyap Chitta{$^{2}$}
\vspace{0.1cm}
\\
{$^{1}$}University of T{\"u}bingen, T{\"u}bingen AI Center \qquad
{$^{2}$}NVIDIA Research \qquad
{$^{3}$}KE:SAI
}

\begin{document}

\twocolumn[{
\maketitle
\vspace{-1.0cm}
\begin{center}
    \vspace{0.35cm}
    \includegraphics[width=1.0\linewidth]{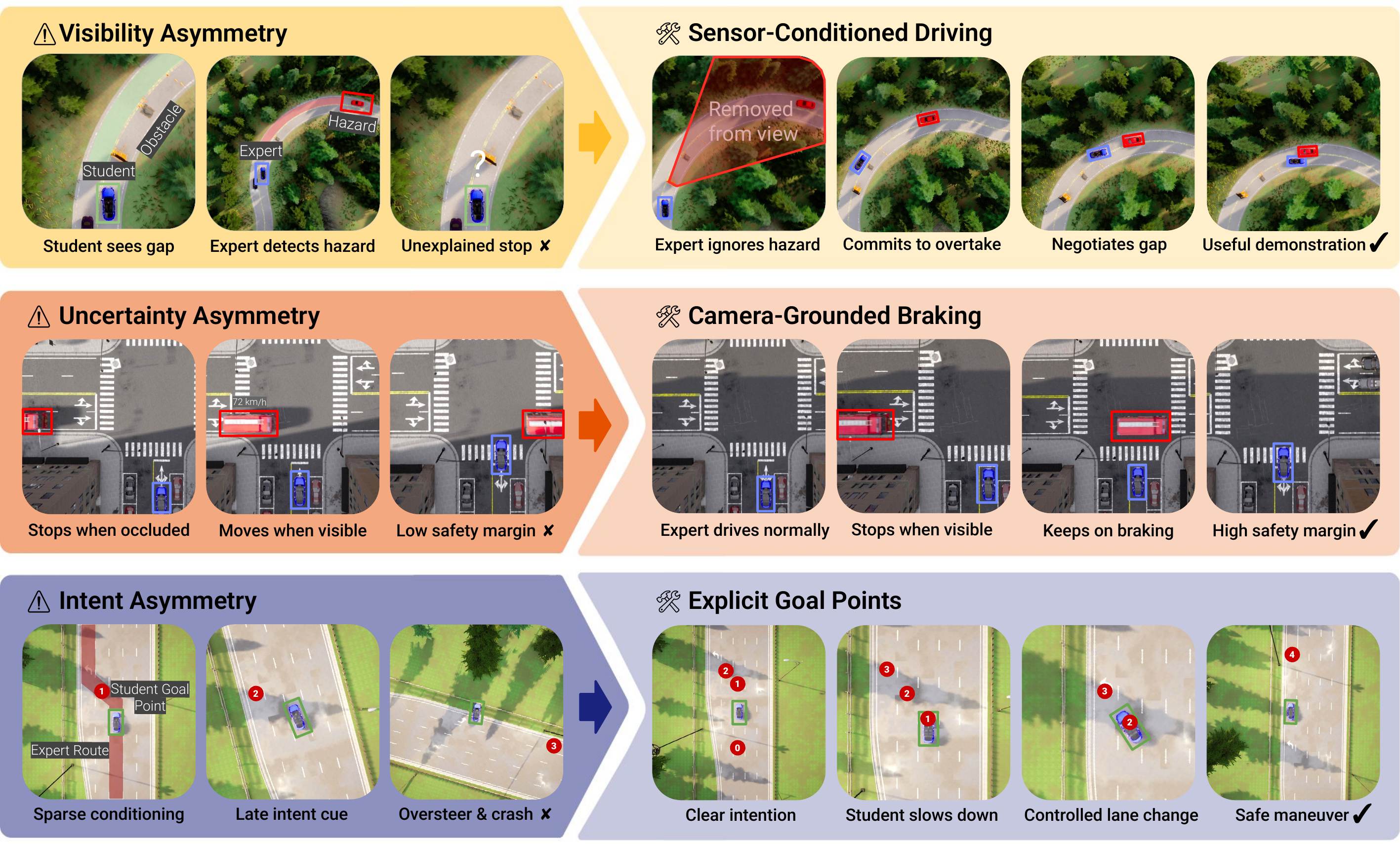}
    \vspace{-0.5cm}
    \captionof{figure}{
\textbf{Performing a task well and teaching it well are not the same.} 
An expert driver (blue bounding box) is most useful when its behavior can be transferred to a student policy (green bounding box) effectively. 
Current expert drivers for CARLA do not fulfill this requirement. We focus on three common asymmetries that hinder effective transfer.
\textit{Visibility asymmetry}: the expert reacts to occluded actors, leading to non-causal and less useful demonstrations.
\textit{Uncertainty asymmetry}: the expert's noiseless state inputs (e.g., accelerations and velocities of other vehicles) lead to successful but dangerous demonstrations.
\textit{Intent asymmetry}: the student's intent is under-specified (as a single target point) making it unaware of complex multi-lane maneuvers.
Our approach reduces expert privileges, enforces sensor-aware demonstrations and redesigns the policy's navigation conditioning, resulting in state-of-the-art closed-loop driving in CARLA.
}
\label{fig:teaser}
\end{center}
\vspace{0.4cm}
}]

\begin{abstract}
Simulators can generate virtually unlimited driving data, yet imitation learning policies in simulation still struggle to achieve robust closed-loop performance.
Motivated by this gap, we empirically study how misalignment between privileged expert demonstrations and sensor-based student observations can limit the effectiveness of imitation learning.
More precisely, experts have significantly higher visibility (e.g., ignoring occlusions) and far lower uncertainty (e.g., knowing other vehicles' actions), making them difficult to imitate reliably.
Furthermore, navigational intent (i.e., the route to follow) is under-specified in student models at test time via only a single target point.
We demonstrate that these asymmetries can measurably limit driving performance in CARLA and offer practical interventions to address them.
After careful modifications to narrow the gaps between expert and student, our TransFuser v6 (TFv6) student policy achieves a new state of the art on all major publicly available CARLA closed-loop benchmarks, reaching 95 DS on Bench2Drive and more than doubling prior performances on Longest6~v2 and Town13.
Additionally, by integrating perception supervision from our dataset into a shared sim-to-real pipeline, we show consistent gains on the NAVSIM and Waymo Vision-Based End-to-End driving benchmarks.
Our code, data, and models are publicly available at \href{https://github.com/kesai-labs/lead}{https://github.com/kesai-labs/lead}.
\end{abstract}

\section{Introduction}
Closed-loop evaluation is essential in robotics, in particular autonomous driving, because strong open-loop prediction does not necessarily translate into robust closed-loop control~\cite{Codevilla2018ECCV, Dauner2024NEURIPS}.
While real-world closed-loop testing remains essential, its cost motivates the widespread use of simulators such as CARLA for benchmarking driving policies~\cite{Dosovitskiy2017CORL}.
Within this setting, a two-stage imitation learning paradigm, often called Learning by Cheating (LBC)~\cite{Chen2019CORL}, has proven highly effective.
First, a privileged expert policy is constructed using ground-truth state information (e.g., the precise map layout), allowing it to plan without perception errors.
The second stage then trains a student policy to reproduce the expert’s actions from sensory observations (e.g., cameras).
A rich body of recent literature supports this methodology~\cite{PDMLITE2024Online, Jia2024NEURIPS, Sima2024ECCV, Renz2025CVPR}.

In initial work following this paradigm, constructing a reliable expert was itself a major challenge, and empirical progress was closely tied to expert quality: early improvements in expert behavior translated directly into better student performance~\cite{Jaeger2021, Chitta2023PAMI}.
As a result, expert design focused primarily on maximizing performance, driven by the expectation that stronger experts would yield stronger students.
Once expert policies achieved strong benchmark performance, most work treated them as fixed sources of supervision, prioritizing model improvements over further expert redesign~\cite{Li2024ECCV, Jia2024NEURIPS}.
This approach was successful when models were the bottleneck, but the landscape has shifted. 
With student policies now plateauing well below expert performance~\cite{Bench2Drive2025Online}, the expert itself deserves renewed attention.

Our work studies, in detail, the generalization gap between expert and student policies.
We identify and analyze several sources of misalignment between expert and student that systematically hinder effective imitation in CARLA, as illustrated in \figref{fig:teaser}.
This motivates a new expert (and corresponding dataset), \textbf{LEAD}, explicitly designed to reduce \underline{L}earner–\underline{E}xpert \underline{A}symmetry in \underline{D}riving.

In parallel to expert misalignment, a second limitation arises from how student policies are conditioned on navigation intent. 
Prior work observes a pronounced \emph{target point bias}, where policies over-emphasize the immediate goal location and under-utilize driving context~\cite{Jaeger2023ICCV}.
This behavior is commonly attributed to weak scene representations~\cite{Jaeger2023ICCV}. 
Our work shows that target point bias persists even when scene representations are made stronger. 
Instead, we attribute it to two additional factors. 
First, driving intent is often insufficiently specified: a single target point does not provide information to disambiguate multi-step maneuvers such as lane changes, making direct goal pursuit a convenient shortcut. 
Second, the geometric target point coordinates are frequently injected late within the decoder of the driving policy, preventing meaningful interaction with perception features from the encoder and increasing the target point's influence relative to them. 
Together, these two factors lead to the most dominant failure modes that negatively impact closed-loop performance of our baseline policy.

Building on these insights, we introduce TransFuser v6 (TFv6), which reduces these asymmetries to achieve substantially improved closed-loop driving. 
When trained on a large, diverse dataset generated by the LEAD expert, TFv6 achieves a new state of the art on Bench2Drive with 95 DS~\cite{Jia2024NEURIPS}, an improvement of 8 DS compared to the previously established best performance. This is significant progress given the gap between the best-performing and fifth-best-performing published methods is approximately 2 DS~\cite{Bench2Drive2025Online}.
Finally, we aggregate the LEAD dataset with real-world datasets into a unified training pipeline, where pre-training on LEAD yields consistent improvements on the NAVSIM and Waymo End-to-End Driving benchmarks~\cite{Dauner2024NEURIPS, Xu2025ARXIV}.
In summary, our contributions span:

\begin{enumerate}[leftmargin=*, itemsep=2pt, topsep=2pt]
    \item \textbf{Reducing asymmetries.}  
    We systematically analyze and reduce learner–expert asymmetries in CARLA by aligning expert supervision with the student’s observable state and strengthening the student's navigation intent, improving the effectiveness of imitation-based driving.

    \item \textbf{Mitigating target point bias.}  We show that the target point bias persists with existing mitigation strategies and identify insufficient as well as poorly integrated intent conditioning as its key contributing factors.

     \item \textbf{A complete training stack.} We release LEAD, a large-scale CARLA dataset and training pipeline enabling state-of-the-art closed-loop driving performance and measurable sim-to-real gains on NAVSIM and Waymo.
\end{enumerate}

\section{Related Work}

\boldparagraph{End-to-End Driving via Imitation Learning}
Imitation learning (IL) trains policies to map sensor observations to actions by imitating expert behavior~\cite{Pomerleau1988NIPS, Bojarski2016ARXIV}. 
Following rapid progress on public benchmarks~\cite{Codevilla2018ICRA, Codevilla2019ICCV, Chen2019CORL, Prakash2021CVPR, Chen2021ICCVa, Chitta2021ICCV, Chen2022CVPRa, Wu2022NEURIPS, Shao2022CORL, Chitta2023PAMI, Shao2023CVPR, Jaeger2023ICCV}, IL has become the standard approach for end-to-end driving~\cite{Chen2024PAMI, Zimmerlin2024ARXIV, Jia2025ICLR, Liao2025CVPR, Renz2025CVPR, Tang2025ICCV, Liu2025NEURIPS}.
While some work relies on human driving data and open-loop evaluation on static datasets~\cite{Hu2023CVPR, Dauner2024NEURIPS, Cao2025CORL, Xu2025ARXIV}, open-loop metrics are known to be unreliable for driving, which is inherently a closed-loop task~\cite{Codevilla2018ECCV, Zhai2023ARXIV, Li2024CVPR, Weng2024CVPR}. 
Closed-loop evaluation is commonly carried out in interactive simulators that allow policies to act, observe, and recover from outcomes of their own decisions~\cite{Wymann2015, Dosovitskiy2017CORL, Shah2017SPRINGER, Li2022PAMI, Yang2023CVPR, Ljungbergh2024ECCV, Zhou2024ARXIV}.
We expand on this line of work in the CARLA simulator~\cite{Dosovitskiy2017CORL}.

\boldparagraph{Learning by Cheating}
Human collection of driving data is costly in simulation, so privileged experts that leverage ground-truth simulator states to generate supervision for an IL student are used ubiquitously---an approach also often called Learning by Cheating~\cite{Chen2019CORL}.
Such experts are either rule-based~\cite{Codevilla2018ICRA, Jaeger2021, Dauner2023CORL, Beißwenger2024Online, Sima2024ECCV} or reinforcement learning–based~\cite{Zhang2021ICCV, Li2024ECCV, Cusumano-Towner2025ICML, Jaeger2025CORL}. 
In this work, we build upon and propose several improvements to the rule-based open-source expert PDM-Lite~\cite{Sima2024ECCV}, which enables controlled analysis of expert–student alignment.
Rather than optimizing expert performance, we focus on making the expert's demonstrations easier to follow for student models, providing downstream benefits.

\boldparagraph{State Asymmetry}  
Student–expert state asymmetry is a long-standing challenge in robotics~\cite{Messikommer2025ARXIV, Weihs2021ARXIV, Warrington2021ARXIV, Walsman2023ICLR}. 
An expert policy that is optimal under full state observability can become suboptimal, even unsafe, when imitated by a student operating under partial observability. 
Adapting the expert and student jointly is a popular approach to mitigate this mismatch~\cite{Messikommer2025ARXIV, Warrington2021ARXIV}. 
While effective in low-dimensional settings, this typically relies on differentiable experts and expensive online interaction. 
In large-scale driving simulation, such assumptions are often impractical. 
Effective expert policies are often rule-based and non-differentiable. 

In our work, we distinguish two practical sources of state asymmetry (\figref{fig:teaser}). 
\textit{Visibility asymmetry} arises when the expert conditions its actions on parts of the scene that fall outside the student’s sensor coverage. 
\textit{Uncertainty asymmetry} stems from differences in state fidelity: experts operate on noise-free quantities (e.g., velocities and accelerations), enabling confident planning with minimal safety margins, while the student must act under substantially higher uncertainty. 
Prior work has shown that both visibility and uncertainty asymmetry can lead to expert demonstrations that are unreliable or unsafe when imitated from raw sensory inputs~\cite{Zimmerlin2024ARXIV}. 
Our work further carefully constrains the expert’s inputs and decision logic to reduce reliance on information that is unobservable or unrealistically precise for the student, improving alignment between expert supervision and sensor-based imitation.

\boldparagraph{Intent Asymmetry}
We refer to \textit{intent asymmetry} as the mismatch between the information provided to the expert and the student for long-horizon decision making: the former acts on a dense route, while the latter must reason about direction from sparse target points, as depicted in \figref{fig:teaser}. 
A natural response is to increase the density of navigation conditioning. 
However, prior attempts that explicitly increased the density of navigation signals have shown limited success~\cite{Zimmerlin2024ARXIV}. 
Our work shows that increasing intent density alone is insufficient; the manner and location of injecting navigation intent into the policy are equally important.

\boldparagraph{Target Point Bias}
Beyond their original role as high-level \textit{navigational guidance}, target points are often implicitly used by E2E driving models as \textit{corrective signals} to recover from covariate shift, steering the vehicle back toward the route after deviations.
This phenomenon is referred to as the target point bias~\cite{Jaeger2023ICCV}. 
We show that this behavior is closely tied to intent asymmetry: when the navigation intent is under-specified and the driving trajectory correlates highly with the target point, a policy naturally learns the shortcut of adjusting the trajectory steering towards the goal. 
By reducing intent asymmetry and restructuring the method by which navigation information is injected into the policy, our approach mitigates this effect, weakening the reliance on target points as implicit corrective signals while maintaining their role as navigational conditioning.

\section{Minimizing Learner-Expert Asymmetry}
\label{sec:MLEA}
In this section, we study the impact of various learner-expert asymmetries in a controlled manner. 
Using a fixed, strong baseline, we systematically reduce learner–expert mismatch by aligning the information used for expert driving and navigation with what is available to the student policy. 
Holding model architecture and dataset scale constant allows us to attribute performance differences to this alignment.

\subsection{Preliminaries}
\label{sec:MLEA-Preliminaries}
We consider the task of navigating through urban scenarios along a predefined route. 
Each route is represented by a sparse sequence of GNSS coordinates, referred to as target points, which provide coarse navigational guidance.

\boldparagraph{Benchmark} 
We conduct evaluation in CARLA 0.9.15 using the Longest6 v2 and Bench2Drive (B2D) benchmarks~\cite{Jia2024NEURIPS, Jaeger2025CORL}. 
Longest6 v2 spans 36 routes of approximately 2\,km each across six towns, while B2D spans 220 routes of approximately 150\,m on all twelve CARLA towns.

\boldparagraph{Metrics}
We use the official CARLA Leaderboard 2.0 evaluation protocol. The primary metric is the Driving Score (DS), which jointly reflects task completion and safety. DS is defined as the product of Route Completion (RC), the fraction of the route completed, and the Infraction Score (IS), a multiplicative penalty factor initialized at 1.0 and decayed with each infraction. All reported results are averaged over three independently trained models with different training seeds. Each model is evaluated once.

\boldparagraph{Baseline} 
Our approach builds on \textit{TransFuser++}~\cite{Jaeger2023ICCV, Zimmerlin2024ARXIV}, the current state of the art on Longest6~v2.
To avoid confusion due to the large number of TransFuser variants, we adopt the versioning nomenclature of~\cite{Jaeger2024Online} and refer to TransFuser++ as \textbf{TFv5}.
TFv5 is an end-to-end (E2E) policy, trained via imitation learning, that produces Bird's Eye View scene tokens from camera and LiDAR inputs through self-attention-based fusion~\cite{Prakash2021CVPR, Chitta2023PAMI}.
For lateral control, a set of learned route queries attend to these scene tokens. 
After a sequence of self- and cross-attention, the queries are further refined by a shallow and low-dimensional Gated Recurrent Unit (GRU)~\cite{chung2014ARXIV}, whose hidden state is initialized with the GNSS target point. 
For longitudinal control, an additional learned query predicts target speed~\cite{Jaeger2023ICCV}.

\subsection{State Alignment}
TFv5 is supervised by the privileged rule-based expert \textit{PDM-Lite}~\cite{Sima2024ECCV}. 
PDM-Lite computes a dense route using A* search over the lane graph and refines this trajectory to avoid static obstacles. 
Potential collisions are forecast using ground-truth 3D bounding boxes and precise dynamic state information, which directly triggers braking behavior. 
While this design is simple and highly effective, it inherently relies on highly privileged information. 
As a result, expert actions may depend on non-observable actors (\emph{visibility asymmetry}) or on motion estimates with unrealistically high precision (\emph{uncertainty asymmetry}). 
While these issues are widely recognized, addressing them in practice remains challenging. 
To this end, we now provide an overview of how we adapt PDM-Lite to better reflect the information available to the student policy.
Further implementation details are provided in supplementary material.

\boldparagraph{Reducing Visibility Asymmetry}
To prevent expert decisions based on unobservable information, we constrain the expert’s planning inputs to signals accessible through the student’s sensors.

\begin{enumerate}
    \item For dynamic actors, we exclude those outside the student’s camera view, such as pedestrians behind the vehicle, while accounting for actor extent and environmental conditions including weather and time of day. This avoids expert reactions to unobservable hazards.
    \item We apply similar constraints to traffic infrastructure. The stopping logic for traffic lights only considers lights within the camera frustum. For traffic signs, explicit speed limit information is not provided to the model and signs are only intermittently visible. We therefore cap the expert’s target speed to the minimum of the posted limit and the typical flow of nearby vehicles, both of which are inferable from local traffic context.
\end{enumerate}

\boldparagraph{Reducing Uncertainty Asymmetry}
To account for uncertainty in student perception, we adjust the expert’s braking behavior.
\begin{enumerate}
    \item Beyond reacting to predicted collision courses, the expert now brakes in the presence of nearby observable hazards, reducing reliance on precise velocity or acceleration estimates that the student cannot reliably infer.
    \item We further adapt expert behavior based on expected perception reliability. Under low-visibility conditions such as nighttime or heavy rain, we reduce the expert’s driving speed to reflect decreased perceptual confidence.
    \item During unprotected turns at junctions, we enlarge the bounding boxes of oncoming actors for collision checks, encouraging safety decisions that rely on conservative spatial margins rather than precise motion prediction.
\end{enumerate}

\noindent
Using this state-aligned expert LEAD, we collect data with identical spatial coverage and scale to the original dataset. As shown in \tabref{tab:state_alignment_core}, training TFv5 on LEAD improves the Driving Score on Longest6 v2 by +11 points and on Bench2Drive by +1.37 points, while the expert’s own performance remains the same (\tabref{tab:carla}).

\begin{table}[H]
\centering
\small
\begin{tabular}{l | c c}
\toprule
\textbf{TFv5 trained with...} & \textbf{Longest6 v2 DS} $\uparrow$ & \textbf{B2D DS} $\uparrow$ \\
\midrule
PDM-Lite dataset~\cite{Sima2024ECCV} & 22.51 \pmsd{4.42} & 83.56 \pmsd{0.32} \\
LEAD dataset (Ours) & \textbf{34.05} \pmsd{1.50} & \textbf{84.94} \pmsd{0.50} \\
\bottomrule
\end{tabular}
\caption{\textbf{Effect of State Alignment}.}
\label{tab:state_alignment_core}
\vspace{-0.2cm}
\end{table}

\subsection{Intent Alignment}
After improving the quality of driving demonstrations, we observe two failure modes remain particularly prominent:

\begin{enumerate}
    \item When the active target point is distant or placed in an unusual position relative to the ego vehicle, such as behind it in a roundabout, trajectory prediction breaks down entirely, producing disjoint and unusable planning outputs.
    \item When the target lies on an adjacent lane, the policy becomes goal-fixated and steers aggressively toward it, ignoring static and dynamic hazards in the surrounding traffic context. The same behavior can be observed when the target point is inside static obstacles.
\end{enumerate}

\noindent
Notably, these failures occur even under clear weather, open road layouts, and low traffic density, where perception and expert supervision are unambiguous.
This points us to limitations in the model rather than the expert, specifically, under-specified intent conditioning and the target point bias. 
We address these via architectural changes, and call the resulting model TFv6. 
Following~\cite{Renz2025CVPR, Gerstenecker2025ARXIV}, we remove the Gated-Recurrent-Unit-based (GRU-based) refinement stage after the planning queries attend to BEV tokens and represent target point as an explicit token alongside the BEV tokens.  
Before embedding, the target point is normalized to $[-1, 1]$ using the training dataset statistics.
\tabref{tab:gru} summarizes the effects of this change, showing improvements of $+6$ DS on Longest6 v2 and $+2$ DS on Bench2Drive. 
The improvement is driven by a marked reduction of the first failure mode, in which the planner previously failed to produce a coherent trajectory on several Longest6 v2 routes when target points were distant or unusually positioned. 
Furthermore, we observe generally that the policy is less sensitive to the exact location of the target point.

\begin{table}[h!]
\centering
\small
\begin{tabular}{l | c c}
\toprule
\textbf{Path planning...} & \textbf{Longest6 v2 DS} $\uparrow$ & \textbf{B2D DS} $\uparrow$ \\
\midrule
With GRU & 34.05 \pmsd{1.50} & 84.94 \pmsd{0.50} \\
Without GRU & \textbf{40.70} \pmsd{2.86} & \textbf{87.26} \pmsd{0.47} \\
\bottomrule
\end{tabular}
\caption{\textbf{Effect of Removing GRU}.}
\label{tab:gru}
\vspace{-0.2cm}
\end{table}

\noindent
For the second failure mode, beyond poorly integrated target points, we hypothesize that the navigation signal itself is too sparse. 
To increase the density of navigation conditioning, we replace single-target conditioning with a compact three-point route representation consisting of the previous, current, and future targets. 
We further reduce the distance threshold at which the current target point is switched out, causing future targets to become relevant earlier and provide stronger supervision to the policy during training. 
For example, when the current target point is only 2-3 meters away, it provides insufficient information about the long-term path to be taken, for which the policy can now rely on the future target point. 
Together, these changes substantially improve closed-loop performance, as shown in \tabref{tab:intent_alignment_core}.
Implementation details and qualitative visualizations are provided in the supplementary material. 

\begin{table}[h!]
\centering
\small
\begin{tabular}{l | c c}
\toprule
\textbf{TFv6 trained with...} & \textbf{Longest6 v2 DS} $\uparrow$ & \textbf{B2D DS} $\uparrow$ \\
\midrule
With one TP & 40.70 \pmsd{2.86} & 87.26 \pmsd{0.47} \\
With three TPs & \textbf{42.13} \pmsd{0.75} & \textbf{89.29} \pmsd{0.45} \\
\bottomrule
\end{tabular}
\caption{\textbf{Effect of Multiple Target Points (TPs)}.}
\label{tab:intent_alignment_core}
\vspace{-0.2cm}
\end{table} 

\subsection{Discussion}

\begin{figure}[t!]
    \centering
    \includegraphics[width=1.0\columnwidth]{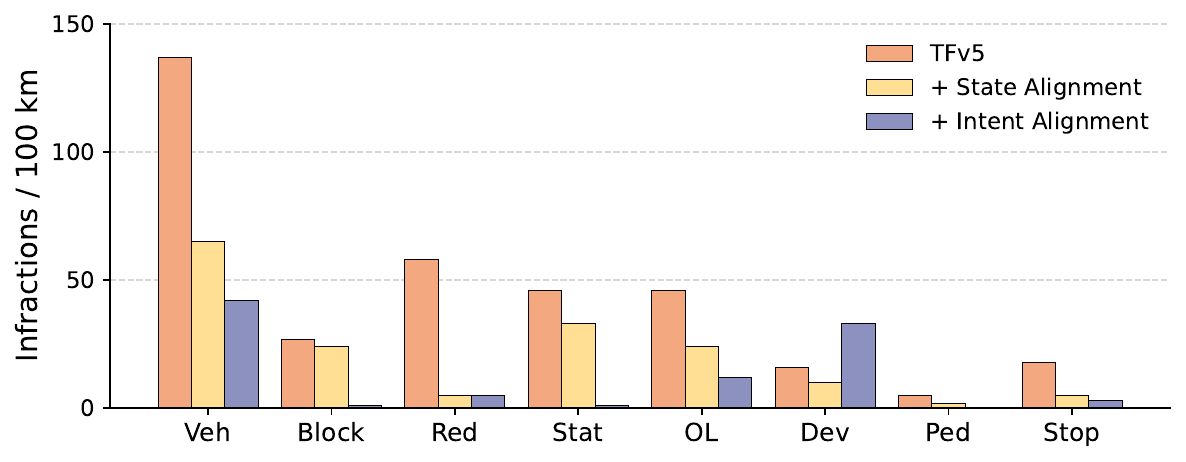}
    \caption{\textbf{Effect of Aligned Supervision and Conditioning.} Infractions counted per 100km, lower is better. \textit{Stat}: Collisions with Layout; \textit{Ped}: Collisions with Pedestrian; \textit{Veh}: Collision Vehicle; \textit{OL}: Outside Lane; \textit{Red}: Red Light; \textit{Dev}: Route Deviation; \textit{SI}: Stop Infraction; \textit{Block}: Vehicle Blocked.}
    \label{fig:submetrics}
\end{figure}

\figref{fig:submetrics} summarizes how state and intent alignment contribute to infraction counts. 
While infractions in general decrease with each improvement, the weakened target point bias, achieved through intent alignment, leads to an increase in route deviation, since the model no longer aggressively snaps back toward the target points after getting off route.

\boldparagraph{Late Goal Conditioning as a Bottleneck}
Although the GRU was originally introduced to model temporal dependencies, its role in modern policy architectures is limited. 
The planning queries already integrate spatial and temporal context through stacked self- and cross-attention, making the GRU largely redundant. 
Inserted after a substantially more expressive transformer decoder (six layers at 256 dimensions), the GRU forms a shallow bottleneck with a single recurrent layer and 64 hidden units. 
Rather than facilitating richer interactions between scene context and the target point, this bottleneck constrains the representation and defaults to reinforcing the influence of the stronger available signal, namely the target point.

Furthermore, the GRU conditions only the planned path and steering on the target point, while target speed prediction is handled independently of the target point. 
This design becomes problematic under distribution shift at deployment. 
During training, the policy is exposed exclusively to on-route data, where steering and speed remain implicitly consistent under nominal route-following behavior. 
Once the vehicle deviates from the route, the GRU-induced decoupling leads to errors: steering is strongly biased toward the target point, while speed prediction is no longer calibrated with respect to the resulting steering behavior.

\section{Experiments}
\label{sec:experiments}

Having established in \secref{sec:MLEA} with a controlled ablation study that aligned supervision and conditioning significantly improve closed-loop performance, we now train TFv6 on an expanded version of LEAD. This dataset is larger and more diverse in terms of towns, lighting, weather conditions, sensor configurations, and scenario coverage.
We evaluate across five benchmarks spanning both simulation and real-world driving. 
Our main experiments focus on long-horizon closed-loop evaluation in CARLA, where we also analyze sensor configurations, and show that combining cameras with LiDAR and radar yields the best performance. 
Finally, we show that synthetic pre-training on LEAD consistently improves performance on multiple real-world open-loop benchmarks, indicating that synthetic data can provide transferable benefits beyond simulation.

\subsection{Closed-Loop Benchmarks}
Building on the controlled ablations in~\secref{sec:MLEA}, we now evaluate TFv6 on long-horizon closed-loop benchmarks. 
In addition to Longest6~v2 and B2D, where the test towns are seen during training, we further consider the challenging Town13 validation benchmark, which consists of 20 routes on an unseen town that are on average 12.39 km long, featuring roughly 100 scenarios per route of 38 different types \cite{Zimmerlin2024ARXIV}. 
Unlike the prior two benchmarks, methods are not allowed to train on data collected in that town, therefore testing the ability of methods to generalize to novel environments (akin to ``level 5" autonomy). 
Town13 validation is among the most challenging driving benchmarks. 

\boldparagraph{Metrics}
We adopt the standard CARLA Leaderboard 2.0 metrics from \secref{sec:MLEA-Preliminaries}. 
On long routes, however, the Driving Score (DS) can be counterintuitive: because the Infraction Score (IS) decays exponentially with each violation, agents are sometimes rewarded for stopping early rather than driving further~\cite{Zimmerlin2024ARXIV}. 
We therefore report the Normalized Driving Score (NDS) on Town13, defined as RC $\times$ I, where I is a distance-normalized version of IS~\cite{Zimmerlin2024ARXIV}. 
For Bench2Drive, we additionally report Success Rate (SR), the fraction of infraction-free completions. Similar to \secref{sec:MLEA}, reported results are averaged over three independent random seeds.

\boldparagraph{Training}
We train TFv6 using a dataset collected with the LEAD expert. 
The dataset is larger than the one used in \secref{sec:MLEA}, containing 73 hours of driving instead of 40 hours. 
We train TFv6 with 4 L40S GPUs for roughly one week in mixed-precision \cite{Micikevicius2018ICLR}. 
Further training and data curation details can be found in the supplementary material.

\boldparagraph{Radar Features}
Radar is a common sensor modality in autonomous driving but rarely included in current CARLA benchmarks. 
To further reduce the learner-expert asymmetry, LEAD provides the student policy with four radar units (each with 75 detections per frame). 
As is common in industrial-grade radars, we pre-process the raw radar detections with a light-weight learned module to provide meaningful object-level features for the downstream policy.
The pre-processing details are provided in the supplementary material. As shown in \figref{fig:architecture}, these object-level radar features bypass the sensor fusion encoder and are input directly into the planning decoder as additional context tokens alongside dense BEV and status information.

\begin{figure}[t!]
    \centering
    \includegraphics[width=1.0\linewidth]{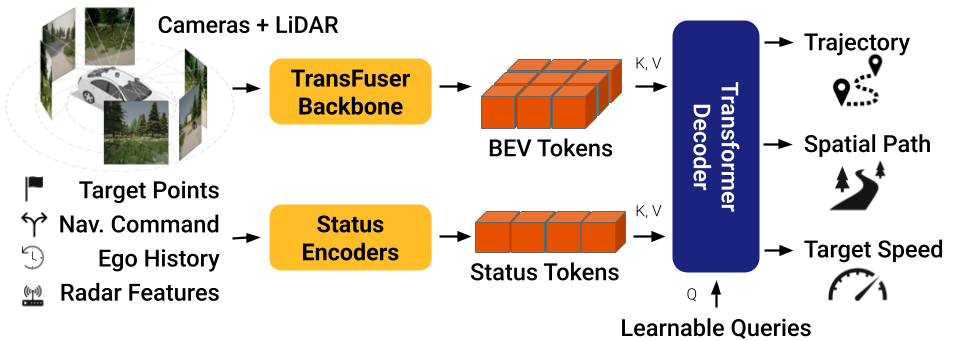}
    \caption{\textbf{TransFuser v6.} The model encodes dense sensor data and sparse status inputs (including multiple target points and per-object radar features) into a unified token space. A transformer decoder then produces a driving plan via learnable queries, removing the GRU bottleneck present in prior TransFuser versions.}
    \label{fig:architecture}
\end{figure}

\boldparagraph{Baselines} 
In addition to TFv5 (\secref{sec:MLEA-Preliminaries}), we include several competitive baselines in our evaluation.
\textit{SimLingo}~\cite{Renz2025CVPR} is a recent vision–language–action model that won the CARLA Leaderboard Challenge 2024, which also builds on PDM-Lite.
\textit{HiP-AD}~\cite{Tang2025ICCV} is a camera-only end-to-end method and the current published state of the art on B2D. 
Note that HiP-AD was trained on a different dataset, which is optimized for B2D.
Finally, \textit{UniAD}~\cite{Hu2023CVPR} is a widely used end-to-end approach that employs sequential auxiliary tasks for perception and planning.

\boldparagraph{Town13 Results}
\tabref{tab:town13} shows the results on the CARLA Town13 benchmark.
Note that we use the best version of our model here, including the radar and LiDAR input modalities, and a RegNet backbone~\cite{radosavovic2020arxiv} (with detailed ablations provided later in \tabref{tab:carla}).
To highlight the impact of generalization, we also evaluate TFv6 trained with data from Town13 (Town13 Train), but shade the results in gray to make it clear that these are only for analysis. 

TFv6 outperforms TFv5 in terms of both DS and NDS, and in both the `Val' and `Train' settings, by large margins.
Interestingly, there is still a significant generalization gap, where TFv6 achieves 14.65 NDS on Town13 Train which drops to 4.04 NDS on Town13 Validation.
This emphasizes the importance of validation benchmarks where methods do not train on any data collected in the validation town.
Qualitatively, TFv6 drives more safely than TFv5 as indicated by its higher IS and I metrics.

\begin{table}[t!]
\small
\setlength{\tabcolsep}{6pt}
    \centering
    \begin{tabular}{l | r | rr | rr}
        \toprule
        \textbf{Method} & ${\mathbf{RC}}$ $\uparrow$ & ${\mathbf{IS}}$ $\uparrow$ & ${\mathbf{DS}}$ $\uparrow$ & $\mathbf{I}$ $\uparrow$ & ${\mathbf{NDS}}$ $\uparrow$ \\
         \midrule
         \multicolumn{6}{c}{{\textbf{`Town13 Val'} -- Town13 withheld during training}} \\
         \midrule
         TFv5~\cite{Zimmerlin2024ARXIV} & \textbf{50.20} & 0.10 & 1.08 & 0.04 & 2.12 \\
         TFv6 (Ours) & 39.70 & \textbf{0.30} & \textbf{3.52} & \textbf{0.22} &  \textbf{4.04}  \\
        \midrule
        \textit{PDM-Lite~\cite{Sima2024ECCV}} & \textit{83.40} & \textit{0.41} & \textit{36.30} & \textit{0.63} & \textit{58.50} \\
        \midrule
        \multicolumn{6}{c}{{\color{lgray} \textbf{`Town13 Train'} -- Trained on all towns}} \\
        \midrule
        {\color{lgray}{UniAD~\cite{Hu2023CVPR}}} & {\color{lgray}{1.42}} & {\color{lgray}{\textbf{0.49}}} & {\color{lgray}{0.23}} & {\color{lgray}{\textbf{0.30}}} & {\color{lgray}{0.00}} \\
        {\color{lgray}{TFv5~\cite{Zimmerlin2024ARXIV}}} & {\color{lgray}{68.53}} & {\color{lgray}{0.04}} & {\color{lgray}{0.96}} & {\color{lgray}{0.07}} & {\color{lgray}{4.94}} \\
        \color{lgray}{TFv6 (Ours)} & \color{lgray}{\textbf{71.82}} & \color{lgray}{0.12} & \color{lgray}{\textbf{5.28}} & \color{lgray}{0.20} & \color{lgray}{\textbf{14.65}} \\
        \bottomrule
    \end{tabular}
    \caption{\textbf{Benchmarking 
    on CARLA Town13.} Results are averaged over three trained models with different training seeds. {\color{lgray}{*Results included for completeness, though this setting is not the recommended default for this benchmark.}}}
    \label{tab:town13}
\end{table}

\begin{table*}[t!]
    \small
    \centering
    \resizebox{\textwidth}{!}{
    \begin{tabular}{l|c|ccc|cc|cc}
        \toprule
        \multirow{2}{*}{\textbf{Method}} & \multirow{2}{*}{\textbf{Backbone}} & \multicolumn{3}{c|}{\textbf{Input Modalities}} & \multicolumn{2}{c|}{\textbf{Bench2Drive}} & \multicolumn{2}{c}{\textbf{Longest6 v2}} \\
        & & Camera FOV & LiDAR & Radar & Driving Score $\uparrow$ & Success Rate $\uparrow$ & Driving Score $\uparrow$ & Route Completion $\uparrow$ \\
        \midrule
        HiP-AD~\cite{Tang2025ICCV} & ResNet-50 & 360° & \xmark & \xmark & 86.8 & 69.1 & 7 & 56 \\
        SimLingo~\cite{Renz2025CVPR} & InternViT-300M & 110° & \xmark & \xmark & 85.1 & 67.2 & 22 & 70 \\
        TFv5~\cite{Zimmerlin2024ARXIV} & RegNetY-032 & 110° & \cmark & \xmark & 83.5 \pmsd{0.3} & 67.3 \pmsd{1.0} & 23 \pmsd{4} & 70 \pmsd{8}\\
        \midrule
        \multirow{6}{*}{TFv6 (Ours)} & \multirow{5}{*}{ResNet-34} & 360°& \xmark & \xmark &  91.6 \pmsd{0.7} & 79.5 \pmsd{2.0} & 43 \pmsd{1} & 85 \pmsd{3} \\
        & & 360° & \cmark & \xmark & 94.7 \pmsd{0.6} & 85.6 \pmsd{0.0} & 52 \pmsd{7}  & 88 \pmsd{5}\\
        & & 360° & \xmark & \cmark & 94.2 \pmsd{0.7} & 85.3 \pmsd{0.9} & 52 \pmsd{1} & 88 \pmsd{2} \\
        & & 360° & \cmark & \cmark & 95.0 \pmsd{0.7} & 84.3 \pmsd{2.1} & 54 \pmsd{5} & 89 \pmsd{3}\\
        & & 140° & \cmark & \cmark & 94.7 \pmsd{0.7} & 82.1 \pmsd{3.6} & 57 \pmsd{3} & \textbf{99 \pmsd{0}} \\
        \cmidrule{2-9}
         & RegNetY-032 & 140° & \cmark & \cmark & \textbf{95.2 \pmsd{0.3}} & \textbf{86.8 \pmsd{0.7}} & \textbf{62 \pmsd{1}} & 91 \pmsd{1}\\
        \midrule
        \textit{PDM-Lite~\cite{Sima2024ECCV}} & - & & - & & \textit{97.0} & \textit{92.3} & \textit{73} & \textit{100} \\
        \textit{LEAD (Ours)} & - & & - & & \textit{96.8} & \textit{96.6} & \textit{73} & \textit{93} \\
        \bottomrule
    \end{tabular}
    }
    \caption{\textbf{CARLA Bench2Drive and Longest6 v2.} TFv6 outperforms all baselines. Our best model uses 140° FOV camera, LiDAR and Radar. Results are averaged over three independently trained models with different training seeds.}
    \label{tab:carla}
\end{table*}

\boldparagraph{Bench2Drive Results}
In \tabref{tab:carla}, we see that scaling the data, backbone, and sensor modalities brings TFv6 within 2 DS of LEAD on B2D, but a 10-point SR gap remains. 
This discrepancy reflects how DS treats short routes: an infraction near the end has little impact when most progress is already secured. 
SR, being all-or-nothing, captures errors that DS discounts. 
This gap indicates the policy is less robust than DS alone suggests.

\boldparagraph{Longest6 v2 Results}
On Longest6 v2, the improvement over TFv5 and SimLingo, both winners of CARLA Challenge 2024, is even more pronounced, with gains of +39 DS and +21 RC. 
While HiP-AD performs competitively on short-route benchmarks, its substantially lower performance on Longest6 v2 illustrates the challenges of long-horizon evaluation. 
This highlights the importance of evaluating driving policies on extended routes, where the likelihood of encountering rare and challenging scenarios increases with route length, exposing weaknesses that short-route benchmarks may miss.
CARLA is currently the only widely used simulator that supports long-form evaluation. 
Most recent benchmarks and simulation frameworks \cite{Gulino2023NEURIPS, Ljungbergh2024ECCV, Karnchanachari2024ICRA, Dauner2024NEURIPS, Zhou2024ARXIV, Kazemkhani2025ICLR, Cao2025CORL, Xu2025ARXIV} are limited to short-form driving due to log-replay or reconstruction-based designs that inherently prevent long-horizon testing. 
Generative approaches \cite{Chitta2024ECCV, Russell2025ARXIV} may offer a path forward in addressing this limitation.

\boldparagraph{Ablation}
\tabref{tab:carla} also shows several ablations on different sensor setups with TFv6. We find that using a wide front camera-setup with 140° FOV with a combination of LiDAR and Radar sensors yields the best results.
Additionally, we show that the RegNetY-032 backbone provides a notable performance improvement compared to the smaller ResNet-34 backbone, consistent with the findings of \cite{Chitta2023PAMI}.
Finally, despite not being optimized for driving performance, LEAD matches the performance of PDM-Lite on Bench2Drive and Longest6 v2, indicating that improved learner–expert alignment does not require sacrificing expert competence.

\subsection{Real-World Data Benchmarks}
To complement our closed-loop CARLA evaluation, we further assess transfer to multiple open-loop real-world benchmarks for end-to-end driving. 

\noindent
\textbf{NAVSIM v1}~\cite{Dauner2024NEURIPS} evaluates 4-second trajectory rollout using multi-view camera inputs, historical states, and discrete commands over a 1.5-second history. Performance is measured using the Predictive Driver Model Score (PDMS), which aggregates collision avoidance, progress, time-to-collision, driving-area compliance and comfort.

\noindent
\textbf{NAVSIM v2}~\cite{Cao2025CORL} extends v1 with a two-stage pseudo-simulation evaluation pipeline designed to better approximate closed-loop behavior. 
In the first stage, trajectories are evaluated from real-world observations using an extended version of PDMS (EPDMS), incorporating additional rule-based and sub-metrics. 
The second stage evaluates the same planner on pre-rendered synthetic observations generated via 3D Gaussian Splatting~\cite{Kerbl2023TOG, Li2025ARXIVb}. 
Stage~2 scores are aggregated using a proximity-based weighting scheme that prioritizes synthetic start states close to the planner’s Stage~1 predicted endpoint, yielding a final score that reflects robustness to small deviations and error recovery.

\noindent
\textbf{WOD-E2E}~\cite{Xu2025ARXIV} consists of driving scenes that over-sample rare, safety-critical events ($<0.03\%$ of daily driving). 
It evaluates 5-second predicted trajectories using the Rater Feedback Score (RFS), a human-annotated metric designed to capture multi-modal, long-tail driving behavior by assigning full expert credit within predefined trust regions and exponentially decaying the score otherwise.

\boldparagraph{Training} 
We collect CARLA data with the camera parameters of target benchmarks, specifically sampled to emphasize dense agent interactions and matching lighting conditions. 
We only use the perception labels from the synthetic CARLA data to avoid the driving-style mismatch between human data collectors and LEAD. 
For the NAVSIM benchmarks, we use the full \textit{navtrain} split and a subset of 100k frames from CARLA for training. 
We train on mixed data in the first 30 epochs, which smoothly excludes the synthetic data, followed by 90 epochs exclusively training on \texttt{navtrain}. 
For WOD-E2E, we subsample 300k frames uniformly for each epoch. 
Here, we pre-train for 30 epochs exclusively on CARLA data, followed by 30 epochs of fine-tuning on the WOD-E2E training split. 
We provide further information about hyper-parameters and data filtering in our supplementary material. 

Since LiDAR and radar data are not available in all the benchmarks, we drop these modalities and replace the LiDAR with a positional encoding, as done in Latent TransFuser (LTF) \cite{Chitta2023PAMI}. We call the resulting method \textbf{LTFv6} to indicate that this version matches the TFv6 architecture.

\begin{table}[t!]
    \centering
    \footnotesize
    \setlength{\tabcolsep}{5.5pt} %
    \begin{tabular}{lccc}
        \toprule
        \textbf{Method} & \textbf{NAVSIM v1} $\uparrow$ & \textbf{NAVSIM v2} $\uparrow$ & \textbf{WOD-E2E} $\uparrow$ \\
        \midrule
        \textit{Ego MLP}~\cite{Dauner2024NEURIPS} & 65.6 & 12.7 & 7.31 \\
        \midrule
        LTF~\cite{Chitta2023PAMI} & 83.8 & 23.1 & - \\
        LTFv6 & 85.4 & 28.3 & 7.51 \\
        + LEAD & 86.4 & 31.4 & 7.76 \\
        \midrule
        \textit{Expert} & \textit{94.5} & \textit{51.3} & \textit{8.10} \\
        \bottomrule
    \end{tabular}
    \caption{\textbf{LTFv6 on Real-World Data.} We report PDMS on the \texttt{navtest} split of NAVSIM v1, EPDMS on the \texttt{navhard} split of NAVSIM v2, and RFS on the validation split of WOD-E2E. Across all benchmarks, our LTFv6 model and the LEAD training strategy achieve consistent performance gains. Results are averaged over three trained models with different training seeds.}
    \vspace{-0.3cm}
    \label{tab:real}
\end{table}

\boldparagraph{Results} 
As shown in~\tabref{tab:real}, while we achieve modest improvements in absolute terms, they are consistent across benchmarks and training setups. 
On NAVSIM, LTFv6 improves in the ego progress, drivable area compliance, and traffic light compliance sub-metrics with minor trade-offs in comfort. 
Our proposed joint pre-training further increases the LTFv6 score across all benchmarks, demonstrating the value of synthetic data despite the presence of distribution shifts. 
Note that ego status can be interpreted as a lower bound while the expert driver upper bounds differ across benchmarks: NAVSIM v1 and WOD-E2E report human driving performance, whereas NAVSIM v2 uses a privileged planner as the upper bound~\cite{Cao2025CORL}.
Further comparisons are provided in the supplementary material.

\section{Conclusion}

We study how misalignment between privileged experts and student policies affects closed-loop driving. 
By reducing visibility, uncertainty, and intent asymmetries, we introduce LEAD, a student-centric expert and dataset designed to produce more transferable supervision. 
This alignment yields substantial closed-loop performance gains even without architectural changes, underscoring expert design as a practical lever in simulation-based imitation learning.

Beyond supervision, our experiments show that policy behavior is further constrained by how navigation intent is specified and integrated. 
Under-specified and late goal-point conditioning amplifies target point bias and promotes shortcut learning, limiting closed-loop robustness in CARLA. 
Restructuring intent conditioning to be denser and injected earlier enables a more balanced interaction between scene understanding and goal following, leading to more stable behavior over long horizons.

Combining aligned supervision with improved intent conditioning, our new model TFv6 achieves state-of-the-art performance across both short- and long-horizon CARLA benchmarks, with particularly strong gains on challenging long-horizon evaluations where errors accumulate over time.
We also provide preliminary evidence of the benefits of co-training on data from simulation and the real world for benchmarks such as NAVSIM and Waymo.
Together, these results suggest that continued progress in simulation-based end-to-end driving will benefit from treating expert design and goal specification as integral, co-evolving components of the learning system.
As end-to-end driving performance in simulation improves, progress increasingly depends on factors beyond model capacity and data scale.

\subsection{Limitations}
We conclude by discussing several limitations of our study and directions for future work. 

\boldparagraph{Remaining Gap to Expert}
Despite narrowing the learner–expert gap, a meaningful difference remains (\tabref{tab:carla}).
We attribute this largely to inherent limitations of behavior cloning, such as compounding errors and the inability to recover from off-route deviations.
Closed-loop training via DAgger or reinforcement learning~\cite{Karkus2025} is a promising direction to further close this gap.

\boldparagraph{Sim-to-Real Transfer} 
Our work investigates perception co-training but does not address planning co-training, where policy behaviors are trained by combining supervision from human driving trajectories and LEAD. 
Moreover, sim-to-real evaluation is currently limited to open-loop and pseudo-closed-loop benchmarks; demonstrating closed-loop real-world benefits remains future work. 

\boldparagraph{Expert Design Scope}
Our study is restricted to a rule-based expert in simulation, and the proposed alignment requires domain-specific tuning.
Whether similar principles apply to learned experts, human demonstrations, or real-world driving remains open.

\subsubsection*{Acknowledgments}
Bernhard Jaeger and Andreas Geiger were supported by the ERC Starting Grant LEGO-3D (850533) and the DFG EXC number 2064/1 - project number 390727645.
Daniel Dauner was supported by the German Federal Ministry for Economic Affairs and Energy within the project NXT GEN AI METHODS (19A23014S).
We thank the International Max Planck Research School for Intelligent Systems (IMPRS-IS) for supporting Bernhard Jaeger and Daniel Dauner.
This research used compute resources at the Tübingen Machine Learning Cloud, DFG FKZ INST 37/1057-1 FUGG as well as the Training Center for Machine Learning (TCML). 
We also thank Lara Pollehn and Simon Gerstenecker for helpful discussions.

{
    \small
    \bibliographystyle{ieeenat_fullname}
    \bibliography{bibliography_long,bibliography,bibliography_custom}
}

\end{document}